\documentclass{article}
\usepackage{ijcai21}

\usepackage{times}
\usepackage{soul}
\usepackage{url}
\usepackage[hidelinks]{hyperref}
\usepackage[small]{caption}
\usepackage{graphicx}
\usepackage{amsmath}
\usepackage{amsthm}
\usepackage{booktabs}
\usepackage{algorithm}
\usepackage{algorithmic}

\usepackage[T1]{fontenc}
\usepackage{newunicodechar}

\usepackage{subfigure}
\usepackage{multirow}
\usepackage{booktabs}
\usepackage{amsmath}
\usepackage{array}
\usepackage{amssymb}
\usepackage{mathtools}
\usepackage{color}

\usepackage{tikz}

\usepackage[normalem]{ulem}

\title{Dependent Multi-Task Learning with Causal Intervention for Image Captioning}

\author{
Wenqing Chen$^{1,2}$
\and
Jidong Tian$^{1,2}$\and
Caoyun Fan$^{1,2}$\and
Hao He$^{1,2}$\footnote{Corresponding authors}\And
Yaohui Jin$^{1,2}$\footnotemark[1]\\
\affiliations
$^1$MoE Key Lab of Artificial Intelligence, AI Institute, Shanghai Jiao Tong University\\
$^2$State Key Lab of Advanced Optical Communication System and Network, \\Shanghai Jiao Tong University\\
\emails
\{wenqingchen, frank92, fcy3649, hehao, jinyh\}@sjtu.edu.cn\\
}

\hypersetup{
  pdfinfo={
   TemplateVersion={IJCAI.2021.0}
  }
}
\begin{document}

\maketitle

\begin{abstract}
  Recent work for image captioning mainly followed an extract-then-generate paradigm, pre-extracting a sequence of object-based features and then formulating image captioning as a single sequence-to-sequence task. Although promising, we observed two problems in generated captions: 1) content inconsistency where models would generate contradicting facts; 2) not informative enough where models would miss parts of important information. From a causal perspective, the reason is that models have captured spurious statistical correlations between visual features and certain expressions (e.g., visual features of "long hair" and "woman"). In this paper, we propose a dependent multi-task learning framework with the causal intervention (DMTCI). Firstly, we involve an intermediate task, bag-of-categories generation, before the final task, image captioning. The intermediate task would help the model better understand the visual features and thus alleviate the content inconsistency problem. Secondly, we apply Pearl's do-calculus on the model, cutting off the link between the visual features and possible confounders and thus letting models focus on the causal visual features. Specifically, the high-frequency concept set is considered as the proxy confounders where the real confounders are inferred in the continuous space. Finally, we use a multi-agent reinforcement learning (MARL) strategy to enable end-to-end training and reduce the inter-task error accumulations. The extensive experiments show that our model outperforms the baseline models and achieves competitive performance with state-of-the-art models.
\end{abstract}

\section{Introduction}

Image captioning is a typical vision-to-language problem, which aims to automatically generate descriptions for images \cite{DBLP:journals/corr/ChenFLVGDZ15}. Recently, a line of studies used an extract-then-generate paradigm with a preliminary task, object detection \cite{DBLP:conf/cvpr/00010BT0GZ18}. These studies first extracted a sequence of object-based region features from the images via a pre-trained detection network, and then used LSTM-based \cite{hochreiter_long_1997} or Transformer-based \cite{vaswani_attention_2017} sequence-to-sequence (seq2seq) models to generate captions. Empirically, the pre-extracted features greatly help models produce high-fidelity captions \cite{DBLP:conf/cvpr/00010BT0GZ18}.

\begin{figure}[t]
\centering
\includegraphics[width=0.45\textwidth]{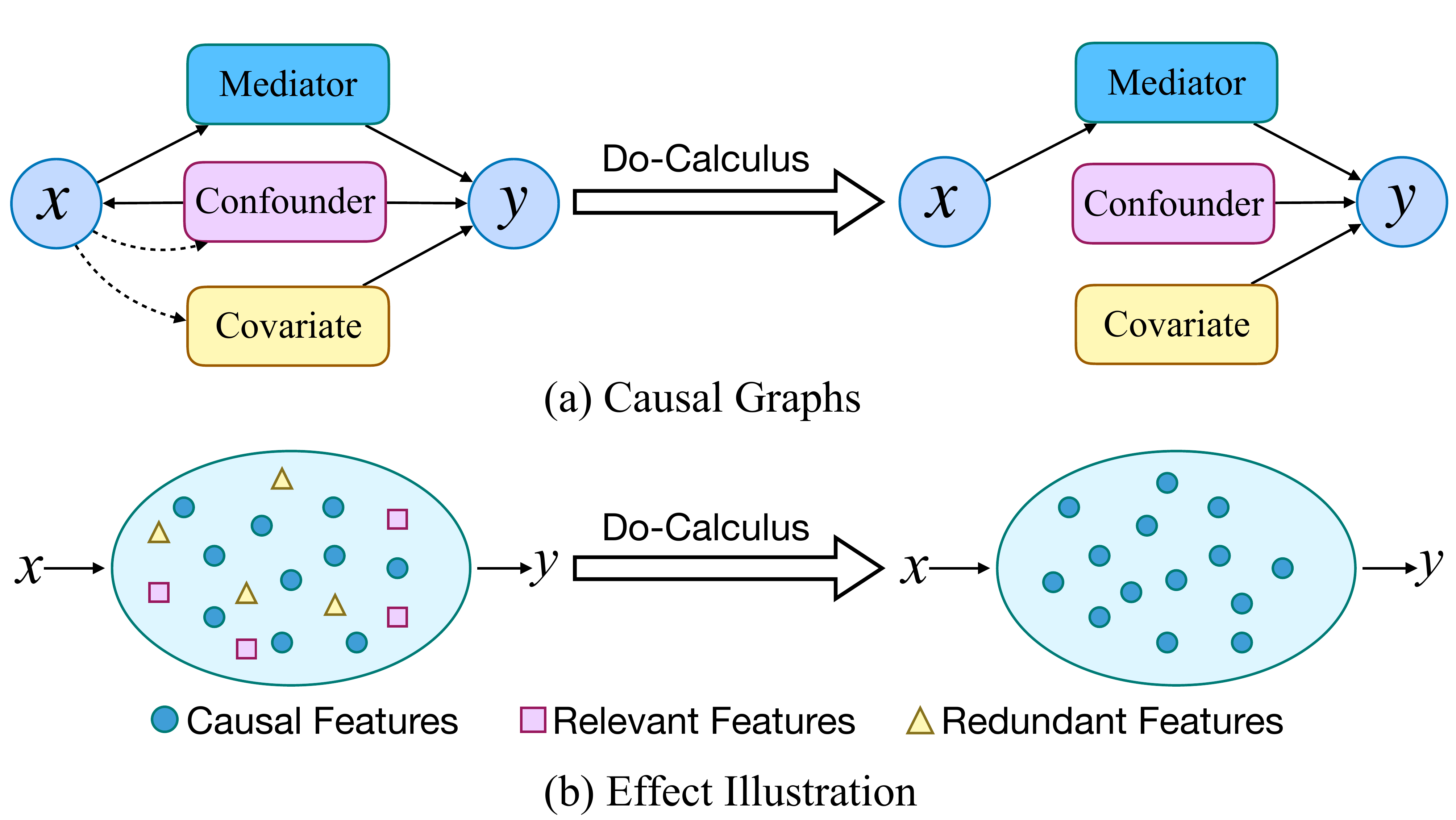}
\caption{The motivation for our work. (a) A typical causal graph including the mediator, the confounder, and the covariate. According to Pearl's do-calculus on the input $\boldsymbol{x}$, the causal graph will be intervened to the right causal graph, which is a robust prediction model. The dashed lines represent the spurious or biased correlations that should not be used for prediction. (b) Illustration of the change of the feature space. }\label{illustration}
\end{figure}

Although the paradigm achieved promising performance, we have observed two problems in the generated captions: 1) content inconsistency; 2) not informative enough. Regarding the first problem, given an image containing a "\emph{man}" with long hair, we found that models often generated "\emph{woman}" instead, which led to a contradicting fact\footnote{The mentioned image and the tested model can be found in Section \ref{sec:case} and Section \ref{sec:settings}, respectively.}. Another example for the second problem is that, given an image with the subject-object-predicate triple ("\emph{man}", "\emph{playing}", "\emph{Wii game}"), we found models ignoring "\emph{Wii game}" and using the quite general words "\emph{video game}" instead. The generated caption would not be judged wrong but not informative enough.

From a causal perspective, the reason is that the models are inclined to build spurious connections between certain attributes of images and high-frequency categories or predicates. For instance, models are possible to capture the spurious statistical correlation between the visual features of "\emph{long hair}" and "\emph{woman}" through the confounder "\emph{female}". Once "\emph{long hair}" is detected, models may incorrectly predict "\emph{man with long hair}" as "\emph{woman}". To further explain this phenomenon, we show a typical causal graph in Figure \ref{illustration}. Let $\boldsymbol{x}$ and $\boldsymbol{y}$ be the given image and one ground-truth caption, respectively. We show the directed acyclic graph (DAG) including the mediator, the confounder, and the covariate. A robust model is supposed to utilize the mediator for prediction, which has causal effects on $\boldsymbol{y}$, rather than utilizing the dashed lines for prediction.

In this paper, we propose a dependent multi-task framework with the causal intervention (DMTCI) to eliminate the spurious correlations. Firstly, we formulate the image caption as a dependent multi-task (DMT) learning problem \cite{DBLP:conf/emnlp/ChenTXHJ20}, which includes two tasks: bag-of-categories (BOC) generation and caption generation. The outputs of the first task are considered as the mediators, which are already annotated in certain datasets like MSCOCO and also utilized by recent work \cite{DBLP:conf/cvpr/WangHZS20a,DBLP:conf/eccv/Li0LZHZWH0WCG20}. Secondly, in terms of the confounders, we are probably to have no access to the real confounders. For example, "\emph{female}" is not annotated. Fortunately, an available approach is using so-called "proxy confounders" \cite{DBLP:conf/nips/LouizosSMSZW17}. Since the spurious correlations problem we observe is that the models over-generating high-frequency categories or predicates, we let the proxy confounders be sampled from a set of concepts including high-frequency predicates and fine-grained object categories. Then we apply the variational inference \cite{DBLP:journals/corr/KingmaW13} to estimate the real confounders and let it sampled from a prior distribution. As a side advantage, our model can also infer covariates from the captions. Finally, we apply a multi-agent reinforcement learning strategy (MARL) to enable end-to-end training and reduce the inter-task error accumulations. An extensive set of experiments demonstrates that our model outperforms the baseline models and achieves competitive performance with those meticulously designed Transformers on the MSCOCO dataset. The main contributions of this work can be summarized as follows:

\begin{itemize}
\item We propose a dependent multi-task framework for image captioning, considering the BOC generation as an intermediate task to alleviate the inconsistency problem.
\item We further introduce a latent de-confounding approach, which reduces the spurious correlations and estimates richer confounders through the variational inference.
\item We cooperate with a multi-agent reinforcement learning approach to better utilize the task dependencies based on our multi-task framework and reduce the inter-task error accumulations.
\end{itemize}

\section{Related Work}

We categorize recent related work on image captioning into three groups according to the different focus of the work.

\textbf{Architectural Innovation.} The first category of work mainly focused on developing novel networks with powerful representation abilities. Some early studies encoded each image as a single feature vector through a pre-trained convolutional neural network (CNN), and then fed it to recurrent neural networks (RNNs) based on long short-term memory units \cite{DBLP:conf/cvpr/VinyalsTBE15,DBLP:conf/cvpr/DonahueHGRVDS15}. Afterward, Xu et al. \shortcite{xu_show_2015} explored the attention mechanisms for image captioning, which could dynamically focus on salient features at each decoding step. Another important line of work cooperated the task "object detection" for image captioning \cite{DBLP:conf/cvpr/KarpathyL15,DBLP:conf/cvpr/FangGISDDGHMPZZ15}. Anderson et al. \shortcite{DBLP:conf/cvpr/00010BT0GZ18} extracted object-level features by a pre-trained Faster R-CNN model, and combined them with the attention mechanism, which achieved significantly higher performance than object-free models. Follow-up researches further captured the relationships among objects by adding self-attention on object-level features \cite{DBLP:conf/iccv/HuangWCW19}, or using graph convolutional networks \cite{DBLP:conf/eccv/YaoPLM18,DBLP:conf/cvpr/YangTZC19}. More recently, Transformer-based \cite{vaswani_attention_2017} models were proposed with spatial position embeddings and better normalization layers \cite{DBLP:conf/cvpr/GuoLZYLL20}, meshed attentions \cite{DBLP:conf/cvpr/CorniaSBC20} or higher-order attentions \cite{DBLP:conf/cvpr/PanYLM20}. It is worthwhile to mention that almost all the state-of-the-art (SOTA) models utilized the self-critical reinforcement (SCRL) strategy \cite{DBLP:conf/cvpr/RennieMMRG17}.

%Moreover, it is worthwhile to mention that almost all SOTA models utilized the self-critical RL strategy proposed by \cite{DBLP:conf/cvpr/RennieMMRG17}, which reduced the train-test discrepancy of caption decoders.

\textbf{Knowledge Enhancement.} The second category of work involved prior knowledge.  Kim et al. \shortcite{DBLP:conf/cvpr/0003COK19} proposed a relationship-based captioning model that considered part-of-speech (POS, i.e. subject-object-predicate categories) tags of captions, and therefore introduced beneficial inductive bias. Shi et al. further explored the semantics conveyed in the subject-object-predicate triples to enhance image caption generation \cite{DBLP:journals/corr/abs-2006-11807}. The subject-object-predicate triples are extracted from captions by a textual scene graph parser introduced in \cite{DBLP:conf/acl-vl/SchusterKCFM15}.

\textbf{Causal Intervention.} The third category of work is quite related to our work, which mainly focused on reducing the spurious correlations with causal intervention. Wang et al. used causal intervention to better learn the region features of the object detection task \cite{DBLP:conf/cvpr/WangHZS20a}. When trying to learn the conditional probability between two objects like "\emph{toilet}" and "\emph{person}", they assumed other objects as confounders and learned the objective of $p(\text{\emph{toilet}} | \text{do}(\text{\emph{person}}))$ by cutting off the link between other co-occurred objects and the object "\emph{person}". Yang et al. assumed the dataset as the confounder and adjust it as concepts, such as "\emph{apple}" and "\emph{green}". Then they cut off the link between these concepts and image features.

\section{Methodology}

In this section, we first formulate the image captioning task as a dependent multi-task learning problem in Section \ref{method:formulation}. Then we illustrate our mediation and de-confounding approach in Section \ref{sec:causal_graph} and \ref{sec:deconfounding}, with the model details in Section \ref{method:DMTCI}. Finally, we introduce the multi-agent reinforcement learning approach in Section \ref{method:MARL}.

\subsection{Dependent Multi-Task Learning}\label{method:formulation}

Given the image and one corresponding ground-truth caption $(\boldsymbol{x}, \boldsymbol{y})$, we use a pre-trained Faster R-CNN network \cite{DBLP:conf/cvpr/00010BT0GZ18} to extract a sequence of features $\boldsymbol{R} \in \mathbb{R}^{L_r \times d}$ from $\boldsymbol{x}$ where $L_r$ and $d$ denote the number of the detected objects and the dimension of features, respectively. The pre-trained Faster R-CNN network is fixed during further training. Recent work mainly formulated image captioning as a single-task learning problem, directly maximizing $p(\boldsymbol{y}|\boldsymbol{R})$ while we formulate image captioning as a DMT problem, which maximizes the log-probability:
\begin{equation}\label{eq:formulation}
\log p(\boldsymbol{y}, \boldsymbol{m}|\boldsymbol{R}) = \log p(\boldsymbol{m}|\boldsymbol{R}) + \log p(\boldsymbol{y}|\boldsymbol{R}, \boldsymbol{m})
\end{equation}
where $\boldsymbol{m}$ is an intermediate variable conveying important information to generate final captions. We choose the BOC as $\boldsymbol{m}$ which are already annotated in MSCOCO dataset and also utilized in recent work \cite{DBLP:conf/cvpr/WangHZS20a,DBLP:conf/eccv/Li0LZHZWH0WCG20}.

\subsection{Mediation with DMT}\label{sec:causal_graph}

There is a train-test discrepancy in Equation \ref{eq:formulation}. In the training stage, we can use ground-truth object categories to supervise the learning objective $\log p(\boldsymbol{m}|\boldsymbol{R})$ while in the inference stage, $\boldsymbol{m}$ is not available. To reduce the train-test discrepancy, we include $\widetilde{\boldsymbol{m}}$, the predicted BOC, for further training after pre-training with Equation \ref{eq:formulation}. The learning objective of caption generation becomes:
\begin{equation}\label{eq:mediation}
\begin{aligned}
 p(\boldsymbol{y}|\boldsymbol{R}) &= \sum_{\boldsymbol{m}} p(\boldsymbol{y} | \boldsymbol{R} , \boldsymbol{m}) p(\boldsymbol{m} | \boldsymbol{R})
\\&  ={\mathbb{E}}_{[\widetilde{\boldsymbol{m}} \sim p(\boldsymbol{m} | \boldsymbol{R})]} \left[p(\boldsymbol{y} | \boldsymbol{R} , \widetilde{\boldsymbol{m}})\right] 
\end{aligned}
\end{equation}
where ${\mathbb{E}}$ denotes the expectation operation. Then the multi-task objective is maximizing:
\begin{equation}\label{eq:mt_formulation}
\begin{aligned}
&\;\;\;\; \log p(\boldsymbol{m}|\boldsymbol{R}) + \log p(\boldsymbol{y}|\boldsymbol{R})
 \\&= \log p(\boldsymbol{m}|\boldsymbol{R}) + {\mathbb{E}}_{[\widetilde{\boldsymbol{m}} \sim p(\boldsymbol{m} | \boldsymbol{R})]} \left[\log  p(\boldsymbol{y} | \boldsymbol{R} , \widetilde{\boldsymbol{m}})\right] 
\end{aligned}
\end{equation}
which utilizes the dependency between tasks and meanwhile reduces the train-test discrepancy. In the period of maximum likelihood estimation (MLE), we use the Gumbel-softmax approach \cite{DBLP:conf/iclr/JangGP17} when sampling $\widetilde{\boldsymbol{m}} \sim p(\boldsymbol{m} | \boldsymbol{R})$ which can back-propagate gradients between tasks (details can be found in Appendix \ref{apd_gs}).

We view Equation \ref{eq:mediation} from the perspective of causal intervention. A typical causal graph including $\boldsymbol{m}$ as a mediator variable is shown in Figure \ref{causal_graphs}. The approach of causal intervention relies on the do-calculus \cite{pearl2010on}. Specifically, when applying the $\text{do}(\cdot)$ operation on the input variable $X$, we are trying to intervened $X$ to the observed value $\boldsymbol{x}$ denoted by $\text{do}(X=\boldsymbol{x})$. After the do-calculus, $\boldsymbol{x}$ (represented by $\boldsymbol{R}$ in practice) is not determined by other variables. In other words, all the arrows pointing to $\boldsymbol{x}$ will be cut off while the arrows from $\boldsymbol{x}$ to other variables remain. It means that, for the mediation causal graph, causal intervention will let $p(\boldsymbol{y}| \text{do}(\boldsymbol{R})) = p(\boldsymbol{y}| \boldsymbol{R})$ which does not modify Equation \ref{eq:mediation}.

\subsection{Latent De-Confounding}\label{sec:deconfounding}

\begin{figure}[t]
\centering
\includegraphics[width=0.48\textwidth]{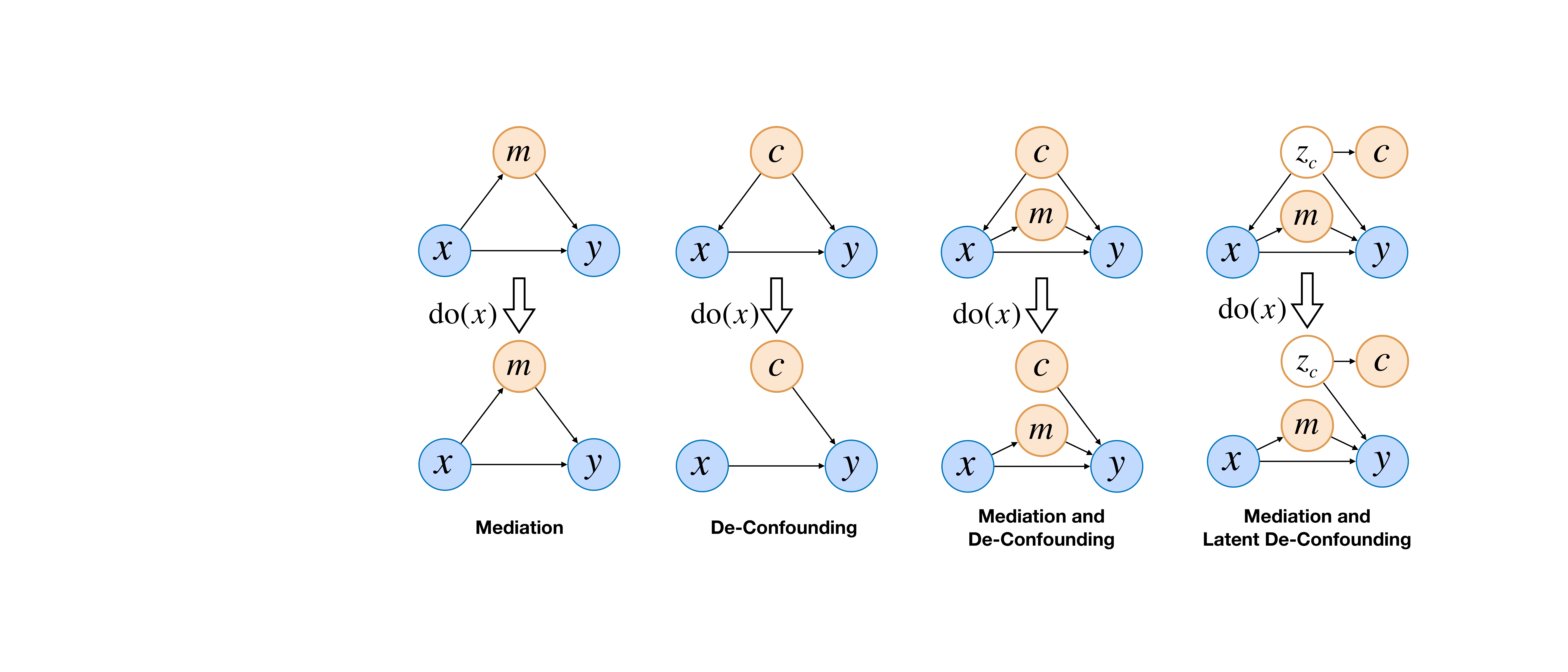}
\caption{Illustration of the change of causal graphs after the do-calculus, which include the mediator $\boldsymbol{m}$, the proxy confounder $\boldsymbol{c}$, and the latent confounder $\boldsymbol{z}_c$.}\label{causal_graphs}
\end{figure}

However, for the de-confounding causal graph, the causal intervention will cut off the arrows pointing to $\boldsymbol{x}$. We denote the proxy confounder and the real confounder by $\boldsymbol{c}$ and $\boldsymbol{z}_z$, respectively. We apply the do-calculus on $\boldsymbol{x}$ and obtain the de-confounding graph shown at the bottom of Figure \ref{causal_graphs}. Firstly, we only consider $\boldsymbol{c}$ and obtain the probability distribution:
\begin{equation}\label{eq:intervention}
\begin{aligned}
&p(\boldsymbol{y}| \text{do}(\boldsymbol{R})) = \sum_{\boldsymbol{c}} p(\boldsymbol{y} | \boldsymbol{R} , \boldsymbol{c}) p(\boldsymbol{c}|\text{do}(\boldsymbol{R})) 
\\&= \sum_{\boldsymbol{c}} p(\boldsymbol{y} | \boldsymbol{R} , \boldsymbol{c}) p(\boldsymbol{c})={\mathbb{E}}_{[\widetilde{\boldsymbol{c}} \sim p(\boldsymbol{c})]} \left[p(\boldsymbol{y} | \boldsymbol{R} , \widetilde{\boldsymbol{c}})\right]
\end{aligned}
\end{equation}
which makes $p(\boldsymbol{y}| \text{do}(\boldsymbol{R})) \neq p(\boldsymbol{y}| \boldsymbol{R})$ as $p(\boldsymbol{c}|\text{do}(\boldsymbol{R})) = p(\boldsymbol{c})$. It means that only considering the mediators $\boldsymbol{m}$ is not optimal during causal intervention. When jointly considering the mediator and the confounder, we do an intervention so that the mediator $\boldsymbol{m}$ is no longer affected by the confounder $\boldsymbol{c}$. Then we have:
\begin{equation}\label{eq:proxyC}
\begin{aligned}
 p(\boldsymbol{y}| \text{do}(\boldsymbol{R})) &= \sum_{\boldsymbol{c}} \sum_{\boldsymbol{m}} p(\boldsymbol{y} | \boldsymbol{R} , \boldsymbol{m}, \boldsymbol{c}) p(\boldsymbol{m} | \boldsymbol{R}) p(\boldsymbol{c})
 \\& = {\mathbb{E}}_{[\widetilde{\boldsymbol{c}} \sim p(\boldsymbol{c}), \widetilde{\boldsymbol{m}} \sim p(\boldsymbol{m}|\boldsymbol{R})]} \left[p(\boldsymbol{y} | \boldsymbol{R} , \widetilde{\boldsymbol{m}}, \widetilde{\boldsymbol{c}})\right]
 \end{aligned}
\end{equation}
according to the mediation-confounding graph in Figure \ref{causal_graphs}.

The next important thing is choosing the proxy confounder $\boldsymbol{c}$ and estimating the real confounder $\boldsymbol{z}_c$. Generally, confounders will have causal effects on $\boldsymbol{x}$ and $\boldsymbol{y}$. For example, "\emph{female}" may have causal effects on the visual features of "\emph{long hair}" in $\boldsymbol{x}$ and the word "\emph{woman}" in $\boldsymbol{y}$. Those models modeling a high posterior probability $q(\text{\emph{female}}|\emph{long hair})$ may incorrectly take a man with long hair as a woman. In this case, "\emph{female}" may be a workable confounder but not be annotated. Thanks to the so-called proxy confounder, we can estimate $\boldsymbol{z}_c$ from proxy confounders $\boldsymbol{c}$ (e.g., "\emph{girl}"). Since we have observed that models over-generated high-frequency expressions, we assume that the proxy confounder $\boldsymbol{c}$ can be sampled from a concept set that includes the high-frequency predicates and fine-grained object categories. We use the summarized predicates and fine-grained categories in \cite{DBLP:journals/corr/abs-2006-11807} as the concept set. The real confounders $\boldsymbol{z}_c$ has causal effects on $\boldsymbol{c}$ and we show the final causal graph of mediation and latent de-confounding in Figure \ref{causal_graphs}.

Estimating $\boldsymbol{z}_c$ requires the variational inference \cite{DBLP:journals/corr/KingmaW13}, where the probability distribution $p(\boldsymbol{z}_c)$ is assumed to be priorly defined, such as a zero-mean unit-variance Gaussian distribution. Taking the modeling of the prior distribution $p(\boldsymbol{c})$ for example, we can use following approximation:
\begin{equation}\label{eq:vae}
\hspace{-1.1mm}
\begin{aligned}
&\log p_{\vartheta, \phi}(\boldsymbol{c}) = \log \int_{\boldsymbol{z}_c} p_{\vartheta}(\boldsymbol{c} |\boldsymbol{z}_c) p(\boldsymbol{z}_c) d \boldsymbol{z}_c
\\&\geq{\mathbb{E}}_{[\boldsymbol{z}_c \sim q_{\phi}(\boldsymbol{z}_c | \boldsymbol{c})]}[\log p_{\vartheta}(\boldsymbol{c} | \boldsymbol{z}_c)]-\text{KL}(q_{\phi}(\boldsymbol{z}_c | \boldsymbol{c}) \| p(\boldsymbol{z}_c))
\end{aligned}
\hspace{-1.5mm}
\end{equation}
where the integral is approximated by maximizing the evidence lower bound (ELBO), and $\vartheta$ and $\phi$ denote the parameters of two networks modeling the prior and the posterior distributions, respectively. And $\text{KL}(\cdot||\cdot)$ denotes the Kullback-Leibler divergence between two distributions. Then, we modify Equation \ref{eq:proxyC} when considering $\boldsymbol{z}_c$ as the confounder:
\begin{equation}\label{eq:Med_LDC}
\begin{aligned}
&\log p_{\theta, \phi}(\boldsymbol{y} |\text{do}(\boldsymbol{R}))= 
\\&{\mathbb{E}}_{[\widetilde{\boldsymbol{c}} \sim p(\boldsymbol{c}),  \boldsymbol{z}_c \sim q_{\phi}(\boldsymbol{z}_c | \widetilde{\boldsymbol{c}}), \widetilde{\boldsymbol{m}} \sim p_{\varphi}(\boldsymbol{m} | \boldsymbol{R})]}[\log p_{\theta}(\boldsymbol{y} | \boldsymbol{R}, \widetilde{\boldsymbol{m}}, \boldsymbol{z}_c, \widetilde{\boldsymbol{c}})]
\end{aligned}
\hspace{-1mm}
\end{equation}
where $\widetilde{\boldsymbol{c}} \sim p(\boldsymbol{c})$ means that we randomly sample concepts from the dataset and $\boldsymbol{z}_c \sim q_{\phi}(\boldsymbol{z}_c | \widetilde{\boldsymbol{c}})$ means that we can infer the actually mattered confounders in the latent space. It is worth noting that we are not designing a network to estimate the posterior distribution $q_{\phi}(\boldsymbol{z}_c | \boldsymbol{y}, \boldsymbol{c})$ because $\boldsymbol{y}$ and $\boldsymbol{c}$ are not strongly coupled. Instead, as introduced in Equation \ref{eq:Med_LDC}, we model $p(\boldsymbol{c})$ in the dataset rather than making a too strong assumption of $\boldsymbol{c}$ paired with $\boldsymbol{y}$.

The final multi-task learning objective is maximizing:
\begin{equation}
\log p_{\varphi}(\boldsymbol{m} |\boldsymbol{R}) + \log p_{\theta, \phi}(\boldsymbol{y} |\text{do}(\boldsymbol{R})) + \log p_{\vartheta, \phi}(\boldsymbol{c})
\end{equation}
where $\log p_{\theta, \phi}(\boldsymbol{y} |\text{do}(\boldsymbol{R}))$ and $\log p_{\vartheta, \phi}(\boldsymbol{c})$ are computed through Equation \ref{eq:Med_LDC} and \ref{eq:vae} respectively.

\subsection{Model Details}\label{method:DMTCI}

To implement the final learning objective, we introduce our model in detail, which can be divided into three main parts: the hierarchical encoder, the BOC generator, and the caption generator.

\textbf{The Hierarchical Encoder}. Given the sequence of object features $\boldsymbol{R}$, we use a hierarchical encoder for further encoding, which consists of multiple layers of multi-head attention network followed with "adding and normalization" layers:
\begin{equation}
\boldsymbol{R}^{(i+1)} = \text{Add\&Norm}(\text{Attn}^{(i)}(\boldsymbol{R}^{(i)} , \boldsymbol{R}^{(i)}, \boldsymbol{R}^{(i)}))
\end{equation}
where $i$ denotes the layer number, $\boldsymbol{R}^{(0)} = \boldsymbol{R}$, and $\text{Attn}(\cdot)$ is the multi-head attention network introduced in \cite{vaswani_attention_2017}. We choose 
the outputs of an intermediate layer $\boldsymbol{R}^{(i_m)}$ to predict the BOC $\boldsymbol{m}$, and the outputs of the final layer $\boldsymbol{R}^{(n)}$ to predict the caption $\boldsymbol{y}$.

\textbf{The BOC Generator}. The first task is a category-level classification task. Fo each image, we convert $\boldsymbol{m}$ to the one-hot label matrix $\boldsymbol{Y}^{m} \in \mathbb{R}^{L_m \times N_m}$ where $L_m$ denotes the number of all categories, and $N_m$ denotes the maximum number of detected objects belongs to each category. We use another attention-based network for prediction:
\begin{equation}
\boldsymbol{H}^{c}=\text{Attn}(\boldsymbol{Q}^{c}, \boldsymbol{R}^{(i_m)}, \boldsymbol{R}^{(i_m)})
\end{equation}
where $\boldsymbol{Q}^{c} \in \mathbb{R}^{L_m \times d}$ denotes the category-specific query matrix in the attention layer and is trainable. Then we use a fully-connected neural network (FCNN) followed with the softmax function along the last dimension to generate the probabilities of object numbers $\widetilde{\boldsymbol{Y}}^{m}$. Then the loss of the first task is computed through cross-entropy:
\begin{equation}
\mathcal{L}^m =-\sum_{j=1}^{L_m} \boldsymbol{Y}^{m}_{j} \log \widetilde{\boldsymbol{y}}^{m}_{j}
\end{equation}
where $\boldsymbol{Y}^{m}_{i}$ and $\widetilde{\boldsymbol{y}}^{m}_{i}$ are two one-hot vectors representing the number of detected objects belong to the $j$-th category in $\boldsymbol{Y}^{m}$ and $\widetilde{\boldsymbol{Y}}^{m}$, respectively. Maximizing the log probability $\log p_{\varphi}(\boldsymbol{m} |\boldsymbol{R})$ is implemented by minimizing $\mathcal{L}^m$.

\textbf{The Caption Generator}. In the second task, we mainly introduce how to implement $p_{\theta}(\boldsymbol{y} | \boldsymbol{R}, \widetilde{\boldsymbol{m}}, \boldsymbol{z}_c, \widetilde{\boldsymbol{c}})$ and $q_{\phi}(\boldsymbol{z}_c | \boldsymbol{c})$. Firstly, we need to re-encode the predicted labels $\widetilde{\boldsymbol{m}}$. Taking the gold label $\boldsymbol{m}$ for example, we involve another two sequences $\boldsymbol{w}^{m}=\{w^{m}_{j}\}$ and $\boldsymbol{id}^{m}=\{id^{m}_j\}$ representing the words and category indexes, respectively. For example, $\boldsymbol{w}^{m} = [\text{person}, \text{dining table}, \text{pizza}, ...]$ and $\boldsymbol{id}^{m}=[1, 2, 3, ...]$, and $\boldsymbol{m}$ marks the number of corresponding categories. We encode the words of each category in $\boldsymbol{w}^{m}$ with a LSTM layer:
\begin{equation}
\boldsymbol{h}^{m}_{j,t} = \text{LSTM}(\text{OH}(w^{m}_{j,t} \boldsymbol{W}_w), \boldsymbol{h}^{m}_{j,t-1})
\end{equation}
where $\text{OH}(\cdot)$ denotes the one-hot operation, $w^{m}_{j,t}$ denotes the $t$-th token in the $j$-th category in $\boldsymbol{w}^{m}$, and $\boldsymbol{W}_w$ denotes the word embedding layer. Assuming $\boldsymbol{H}^{m} = \{\boldsymbol{h}^{m}_{j, T_m}\}$ containing the hidden states at final tokens (at the $T_m$-th step) and $m_j$ representing the occurrence numbers of $j$-th category in the BOC $\boldsymbol{m}$, we fuse three kinds of embeddings as follows:
\begin{equation}
\boldsymbol{e}^{m}_{j} = (\boldsymbol{h}^{m}_{j,T_m} + \text{OH}(id^{m}_j) \boldsymbol{W}_o) \odot \sigma(\text{OH}(m_j) \boldsymbol{W}_m)
\end{equation}
where $\odot$ denotes the element-wise multiplication and $\sigma$ denotes the sigmoid function. $\boldsymbol{W}_o, \boldsymbol{W}_m$ denote parameters of the category embedding and number embedding layers, respectively. In this way, the final embedding matrix $\boldsymbol{E}^{m} = \{\boldsymbol{e}^{m}_{j}\} \in \mathbb{R}^{L_m \times d}$ is category-words-aware and number-aware. Then we use the max-pooling operation to obtain a vector $\boldsymbol{z}_m$ representing the global information for decoding. It is worth noting that we do not further encode $\boldsymbol{E}^{m}$ with position-aware encoders such as LSTM or attention networks, because the BOC does not contain the detailed information. For example, BOC can not indicate whether a person is hitting a ball or holding a ball. Therefore, we only apply the pooling operation and get $\boldsymbol{z}_m$.

Secondly, we need to infer $\boldsymbol{z}_c$ through the network $q_{\phi}(\boldsymbol{z}_c | \boldsymbol{c})$. We extract the high-frequency predicates and fine-grained categories from $\boldsymbol{y}$ and convert them to an unordered set $\boldsymbol{c}$. We embed $\boldsymbol{c}$ to $\boldsymbol{E}^{c} \in \mathbb{R}^{L_c \times d}$ through the word embedding layer and pass it to a two-layer Transformer without the position embedding, and obtain a sequence of hidden states $\boldsymbol{H}^{z} \in \mathbb{R}^{L_c \times d}$. Since $q_{\phi}(\boldsymbol{z}_c | \boldsymbol{c})$ is often approximated to a Gaussian distribution, we use the mean pooling of $\boldsymbol{H}^{z}$ followed with two FCNNs to get the corresponding mean and variance vectors, and then $\boldsymbol{z}_c$ is sampled from this Gaussian distribution.

Thirdly, we need to implement $p_{\theta}(\boldsymbol{y} | \boldsymbol{R}, \widetilde{\boldsymbol{m}}, \boldsymbol{z}_c, \widetilde{\boldsymbol{c}})$ where $\widetilde{\boldsymbol{m}} \sim p_{\varphi}(\boldsymbol{m} |\boldsymbol{R})$, $\boldsymbol{z}_c \sim q_{\phi}(\boldsymbol{z}_c | \widetilde{\boldsymbol{c}})$ and $\widetilde{\boldsymbol{c}} \sim p(\boldsymbol{c})$. After randomly selecting $\widetilde{\boldsymbol{c}}$ from the dataset, we obtain a vector $\boldsymbol{z}_c$ by sampling from $q_{\phi}(\boldsymbol{z}_c | \widetilde{\boldsymbol{c}})$. Since $\widetilde{\boldsymbol{m}}$ has been already encoded to $\boldsymbol{z}_m$, we simply use a top-down attention LSTM network \cite{DBLP:conf/cvpr/00010BT0GZ18} to decode each token:
\begin{equation}
\mathcal{L}^g=-\sum_{t=1}^{T} \log \left(p_{\theta}\left(y_{t} \mid \boldsymbol{y}_{<t}, \boldsymbol{R}^{(n)}, \boldsymbol{z}_m, \boldsymbol{z}_c\right)\right)
\end{equation}
where $y_{t}$ denote the $t$-th token of $\boldsymbol{y}$, and $\boldsymbol{y}_{<t}$ denotes the sequence of tokens preceding $y_{t}$. The latent variable $\boldsymbol{z}_m$ and $\boldsymbol{z}_c$ are concatenated with the hidden states at all steps. Note that our model is also compatible with Transformer-based decoders. But recent empirical work has shown that Transformer-based decoders require meticulous designs, such as meshed attention \cite{DBLP:conf/cvpr/CorniaSBC20} or high-order attention \cite{DBLP:conf/cvpr/PanYLM20}, and greatly increase the time complexity when optimized with reinforcement learning (RL).

\begin{table*}[!htb]
\resizebox{0.9\textwidth}{!}{\begin{tabular}{l|rrrrrr|rrrrrr}
\toprule & \multicolumn{6}{c} { Maximum Likelihood Estimation } & \multicolumn{6}{c} { Reinforcement Learning } \\
\cmidrule { 2 - 13 } & B1 & B4 & ME & RG & CD & SP & B1 & B4 & ME & RG & CD & SP \\ \midrule
SCST & - & 31.3 & 26.0 & 54.3 & 101.3 & - & - & 33.3 & 26.3 & 55.3 & 111.4 & - \\
Up-Down & 77.2 & 36.2 & 27.0 & 56.4 & 113.5 & 20.3 & 79.8 & 36.3 & 27.7 & 56.9 & 120.1 & 21.4 \\
SGAE & 77.6 & 36.9 & 27.7 & 57.2 & 116.7 & 20.9 & 80.8 & 38.4 & 28.4 & 58.6 & 127.8 & 22.1 \\
AoANet & 77.4 & 37.2 & 28.4 & 57.5 & 119.8 & 21.3 & 80.2 & 38.9 & 29.2 & 58.8 & 129.8 & 22.4 \\
Up-Down + VC & - & - & - & - & - & -  & - & 39.5& 29.0 & 59.0 & 130.5 & - \\ 
WeakVRD (MT-I) & 78.1 & \textbf{38.4} & 28.2 & 58.0 & 119.0 & 21.1 & 80.8 & 38.9 & 28.8 & 58.7 & 129.6 & 22.3 \\
$\mathcal{M}^{2}$ Transformer & - & - & - & - & - & - & 80.8 & 39.1 & 29.2 & 58.6 & 131.2 & 22.6\\
XLAN $\dagger$ &  78.3 & 38.0 & 28.7 & 57.9 & 121.0 & 21.8 & 80.6 & 39.4 & \textbf{29.5} & 59.2 & 131.1 & \textbf{23.1} \\
\midrule
TransLSTM (Baseline) & 77.1 & 37.1 & 28.2 & 57.3 & 117.3 & 21.2 & 80.2 & 38.6 & 28.5 & 58.3 & 128.7 & 22.2 \\
DMTCI (Avg.) & 78.5 & 38.0& 28.5 & 58.1 & 120.2 & 22.0 &  81.3 & 39.7 & 29.1 & 59.1 & 131.7 & 22.9 \\
DMTCI + Trained Selector (20) & \textbf{79.1} & 38.2 & \textbf{28.8} & \textbf{58.2} & \textbf{123.7} & \textbf{22.1} &  \textbf{81.3} & \textbf{39.7} & 29.2 & \textbf{59.2} & \textbf{132.0} & 23.0\\
\midrule
DMTCI + Gold Categories (Avg.) & 79.3 & 38.2& 29.0 & 58.4 & 122.9 & 22.2 &  81.9 & 40.3 & 29.5 & 59.6 & 135.0 & 23.1 \\
DMTCI + Optimum Selector (20) & 82.6 & 45.0 & 31.1 & 62.2 & 141.9 & 23.9 &  82.7 & 42.5 & 30.0 & 61.0 & 139.0 & 23.3 \\
XLAN  $\ddag$ &  78.0 & 38.2 & 28.8 & 58.0 & 122.0 & 21.9 & 80.8 & 39.5 & 29.5 & 59.2 & 132.0 & 23.4 \\
\bottomrule
\end{tabular}}
\centering\caption{Single-model performances on the MSCOCO dataset (Karpathy split) in both MLE and RL period. B1, B4, ME, RG, CD, and SP denote BLEU1, BLEU4, METEOR, ROUGE, CIDEr-D, and SPICE, respectively. Most results of SOTA models are cited except that models tagged by $\dagger$ are reproduced with the officially released codes. We also show the officially reported results of XLAN marked with $\ddag$.}\label{tab:mainresults}
\end{table*}

\subsection{Multi-Agent Reinforcement Learning}\label{method:MARL}

RL is commonly used in the task of image captioning to reduce the train-test discrepancy \cite{ranzato2015sequence}. Since we formulate the image captioning as the dependent multi-task problem, the inter-task train-test discrepancy also needs to be tackled. During the MLE optimization period, Gumbel-softmax approximation is used but will improve the gradient variance because of the temperature parameter \cite{DBLP:conf/iclr/NieNP19}. Therefore, we use a \emph{multi-agent reinforcement learning} strategy to train the model end-to-end based on the SCRL strategy on the two tasks alternatively.

\textbf{SCRL for the BOC generator}. The Monte-Carlo sampling is used to generate $\widetilde{\boldsymbol{m}}$ and the arg-max operation to generate the baseline $\overline{\boldsymbol{m}}$. The gradient of the BOC generator can be expressed by:
\begin{equation}
\nabla_{\varphi} \mathcal{L}(\varphi) \approx-\left(r_1(\widetilde{\boldsymbol{m}})-r_1(\overline{\boldsymbol{m}})\right) \nabla_{\varphi} \log p_{\varphi}(\widetilde{\boldsymbol{m}} |\boldsymbol{R})
\end{equation}
where $r_1(\widetilde{\boldsymbol{m}})$ is the reward provided by both two tasks. After feeding $\widetilde{\boldsymbol{m}}$ to the caption generator, we obtain $\overline{\boldsymbol{y}}_{\widetilde{\boldsymbol{m}}}$ through greedy decoding and compute the reward $r_1(\widetilde{\boldsymbol{m}})$ by:
\begin{equation}
r_1(\widetilde{\boldsymbol{m}}) = \lambda_1 s_1(\widetilde{\boldsymbol{m}}, \boldsymbol{m}) + \lambda_2 s_2(\overline{\boldsymbol{y}}_{\widetilde{\boldsymbol{m}}}, \boldsymbol{y})
\end{equation}
where $\lambda_1$ and $\lambda_2$ are the hyperparameters to balance the two rewards, and $s_1(\cdot)$ denotes the score function to match two BOCs, and $s_2(\cdot)$ denotes the CIDEr-D score function \cite{DBLP:conf/cvpr/VedantamZP15}.

\textbf{SCRL for the caption generator}. Fed with the baseline BOC $\overline{\boldsymbol{m}}$ and a sampled latent confounder $\boldsymbol{z}_c \sim p(\boldsymbol{z}_c)$, the caption generator can obtain $\widetilde{\boldsymbol{y}}$ by the Monte-Carlo sampling and $\overline{\boldsymbol{y}}$ by the greedy decoding. We use the SCRL strategy with the gradients computed as follows:
\begin{equation}
\nabla_{\theta} \mathcal{L}(\theta) \approx-\left(r_2(\widetilde{\boldsymbol{y}})-r_2(\overline{\boldsymbol{y}})\right) \nabla_{\theta} \log p_{\theta}(\widetilde{\boldsymbol{y}} | \boldsymbol{R}, \overline{\boldsymbol{m}}, \boldsymbol{z}_c)
\end{equation}
where $r_2(\widetilde{\boldsymbol{y}}) = s_2(\widetilde{\boldsymbol{y}}, \boldsymbol{y})$. Furthermore, we adopt the max-max framework \cite{DBLP:conf/aaai/XiaoWHJ20} which maximizes the rewards of the two agents alternatively.

\section{Experiments}
\subsection{Datasets and Settings}\label{sec:settings}

\textbf{Datasets}. We experiment on the MSCOCO\footnote{http://cocodataset.org/} dataset, which is the most popular dataset for image captioning. The original dataset contains about 82,783 training images and 40,504 validation images. Following most of the previous work, we first evaluate our model on the “Karpathy” data split \cite{DBLP:conf/cvpr/KarpathyL15} with 5,000 images for validation, 5,000 images for testing, and the rest for training. Then we evaluate our model on the on-line COCO test server. We set the maximum decoding length to 16 which equals the maximum length of $95\%$ of captions. We keep the words that occur no less than 5 times, resulting in a vocabulary of 10,369 words.

\textbf{Evaluation and Settings}. We apply five standard metrics for evaluation: CIDEr-D \cite{DBLP:conf/cvpr/VedantamZP15}, BLEU \cite{papineni2002bleu}, METROT \cite{DBLP:conf/acl/BanerjeeL05}, ROUGE \cite{lin2004rouge:}, and SPICE \cite{DBLP:conf/eccv/AndersonFJG16}. All the metrics are computed with the publicly released code.\footnote{https://github.com/tylin/coco-caption}

\textbf{Compared Models.} We compare our models with several models including SCST \cite{DBLP:conf/cvpr/RennieMMRG17}, Up-Down \cite{DBLP:conf/cvpr/00010BT0GZ18}, AoANet \cite{DBLP:conf/iccv/HuangWCW19}, Up-Down+VC \cite{DBLP:conf/cvpr/WangHZS20a}, WeakVRD \cite{DBLP:journals/corr/abs-2006-11807}, $\mathcal{M}^{2}$ Transformer \cite{DBLP:conf/cvpr/CorniaSBC20}, and XLAN \cite{DBLP:conf/cvpr/PanYLM20}. Among them, XLAN is the recent SOTA with an advanced architecture, and Up-Down+VC is related to our work which uses causal intervention to reduce spurious correlations between objects. Also, we compare with the baseline model, TransLSTM, consisting of a Transformer encoder and an up-down LSTM attention decoder. It is worth noting that one may find that a recent model OSCAR \cite{DBLP:conf/eccv/Li0LZHZWH0WCG20} greatly outperforms the models mentioned above, but we want to clarify that OSCAR make two unfair settings: 1) using more vision-language datasets for pre-training, including Conceptual Captions, SBU captions, Flicker30k, and GQA \cite{DBLP:conf/eccv/Li0LZHZWH0WCG20}; 2) utilizing the ground-truth rather than generated object categories in the inference phase.

\textbf{Hyperparameters.} We use the pre-trained Faster R-CNN model \cite{DBLP:conf/cvpr/00010BT0GZ18} to represent each image as an adaptive sequence of feature vectors $\boldsymbol{R}$ where the number of detected objects $n \in[10,100]$. The number of encoder layers is $4$ and we use the output of the second layer to predict BOC. The dimension of each feature vector $v_{i}$ is 2,048 and then projected to 1,024. The dimensions of the word embedding layer, attention layers, and the LSTM-based decoder are all set to 1,024. The learning rate is initialized to $0.0001$ and decreased by half when the CIDEr-D score does not increase in 2 epochs, with the minimum learning rate set to $5\cdot 10^{-6}$. The batch size is set to $50$. The model is firstly optimized with MLE for $30$ epochs (Gumbel sampling $\widetilde{\boldsymbol{m}}$ after $15$ epochs), and then optimized with MARL for another $35$ epochs. The special setting of our model is that we can sample different $\boldsymbol{z}_c$ to generate multiple captions. Therefore, we train an image-text selector to select the best of $20$ generated captions for each image (details in Appendix \ref{apd_selector}).

\begin{table}[!tb]
\resizebox{0.48\textwidth}{!}{\begin{tabular}{l|rrrrrrrr}
\toprule Model &  \multicolumn{2}{c} { B4 } & \multicolumn{2}{c} { ME } & \multicolumn{2}{c} { RG } & \multicolumn{2}{c} { CD } \\
\midrule Metric  & $\mathrm{c} 5$ & $\mathrm{c} 40$ & $\mathrm{c} 5$ & $\mathrm{c} 40$ & $\mathrm{c} 5$ & $\mathrm{c} 40$ & $\mathrm{c} 5$ & $\mathrm{c} 40$ \\
\midrule SCST  & 35.2 & 64.5 & 27.0 & 35.5 & 56.3 & 70.7 & 114.7 & 116.0 \\
Up-Down & 36.9 & 68.5 & 27.6 & 36.7 & 57.1 & 72.4 & 117.9 & 120.5 \\
SGAE & 38.5 & 69.7 & 28.2 & 37.2 & 58.6 & 73.6 & 123.8 & 126.5 \\
AoANet &  37.3 & 68.1 & 28.3 & 37.2 & 57.9 & 72.8 & 124.0 & 126.2 \\
Up-Down + VC &  37.8 & 69.1 & 28.5 & 37.6 & 58.2 & 73.3 & 124.1 & 126.2 \\
WeakVRD (MT-I)& \textbf{38.6} & \textbf{70.1} & 28.6 & 37.8 & \textbf{58.8} & \textbf{74.5} & 125.1 & 126.7 \\
XLAN $\dagger$  & 38.3 & 69.2 & 28.5 & 38.2 & 58.4 & 74.0 & 125.0 & 127.1 \\\midrule 
DMTCI  &38.5 & 70.0 & \textbf{29.0} & \textbf{38.4} & 58.7 & 74.2 & \textbf{125.1} & \textbf{127.6}  \\
\bottomrule
\end{tabular}}
\centering\caption{The single-model performance on COCO online test server.}\label{tab:onlineresults}
\end{table}

\begin{table}[!thb]
\resizebox{0.48\textwidth}{!}{\begin{tabular}{c|l|rrrrrr}
\toprule Period &Ablated Models & B1 & B4 & ME & RG & CD & SP \\ \midrule
\multirow{6}{*}{MLE} &TransLSTM & 77.1 & 37.1 & 28.2 & 57.3 & 117.3 & 21.2  \\
&Pipeline & 77.1 & 36.7 & 27.9 & 56.9 & 116.1 & 20.8  \\
&DMTCI-M & 77.3 & 37.8 & 28.3 & 57.8 & 119.5 & 21.8  \\
&DMTCI-LDC& 77.4 & 37.2 & 28.1 & 57.5 & 118.8 & 21.6 \\
&DMTCI-Full (Avg.) & 78.5 & 38.0& 28.5 & 58.1 & 120.2 & 22.0\\
&DMTCI-Full (Best) & \textbf{79.1} & \textbf{38.2} & \textbf{28.8} & \textbf{58.2} & \textbf{123.7} & \textbf{22.1} \\\midrule
\multirow{6}{*}{RL} &TransLSTM & 80.2 & 38.6 & 28.5 & 58.3 & 128.7 & 22.2  \\
&Pipeline & 79.9 & 38.1 & 28.3 & 58.0 & 126.6 & 22.1 \\
&DMTCI-M & 81.1 & 39.3 & 28.9 & 59.0 & 131.3 & 22.7  \\
&DMTCI-LDC& 80.7 & 38.7 & 28.7 & 58.7 & 130.1 & 22.3  \\
&DMTCI-Full (Avg.)& 81.3 & 39.7 & 29.1 & 59.1 & 131.7 & 22.9  \\
&DMTCI-Full (Best)&  \textbf{81.3} & \textbf{39.7} & \textbf{29.2} & \textbf{59.2} & \textbf{132.0} & \textbf{23.0}  \\
\bottomrule
\end{tabular}}
\centering\caption{The performances of ablated models on the Karpathy split in both MLE and RL periods.}\label{tab:ablation}
\end{table}

\subsection{Main Results}\label{main_results}

Table \ref{tab:mainresults} presents the performance on the Karpathy test split of our model as well as the compared models. As shown, our model DMTCI achieves average CIDEr-D scores of $120.2$ and $131.7$ and outperforms the baseline TransLSTM by $2.9$ and $3.0$ points in MLE and RL periods, respectively. When compared with the related model with causal intervention, "Up-Down+VC", our model outperforms it by $1.2$ points on the CIDEr-D score. Furthermore, when combining with the trained selector, our model outperforms TransLSTM by $6.4$ and $3.3$ points in MLE and RL periods, respectively. And our model achieves competitive results with the recent SOTA model XLAN, without using meticulously designed attention mechanisms. It is worth noting that we also reproduce XLAN with the officially released codes, because in recent work, a slightly tricky method is to obtain an ensemble model by training the model multiple times. Thus, the single-model performance is reported under the best trial.

If we have the gold BOC, the averaged CIDEr-D score can be further improved to $122.9$ and $135.2$ in MLE and RL periods respectively. It suggests that future work can focus on improving the performance of the BOC generator. Moreover, if we have an optimal selector, we can further improve the CIDEr-D score to $141.9$ in the MLE period even without RL, which can be considered as the upper bound of our model. And RL slightly decreases the upper bound, which means that optimizing with RL drops some useful patterns.

Table \ref{tab:onlineresults} presents the single-model results of the models on the online COCO test server. Compared with recent SOTA models which have reported single-model performance, DMTCI achieves competitive performance and achieves the best CIDEr-D score with 40 reference captions.

\begin{figure}[ht]
\centering
\includegraphics[width=0.48\textwidth]{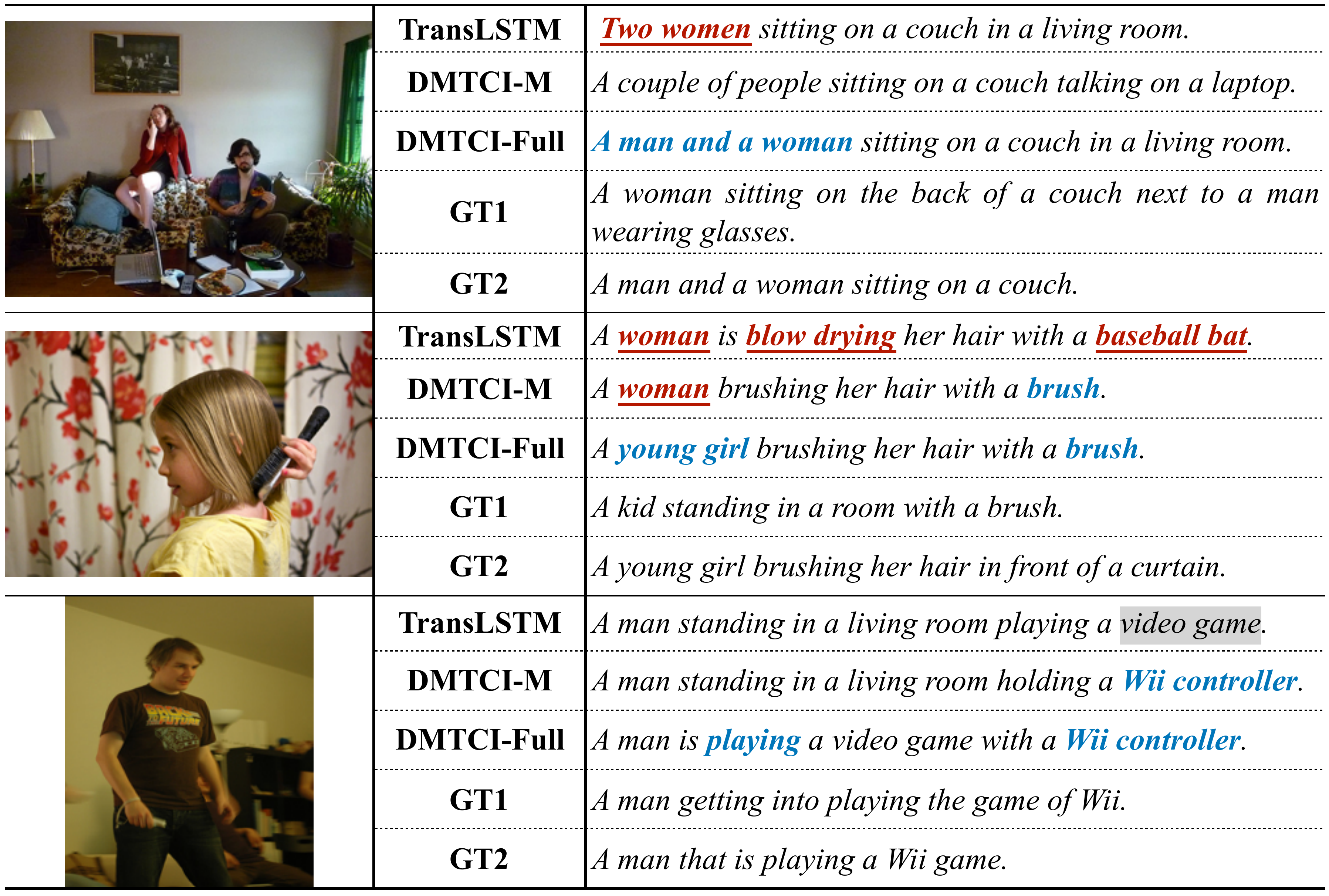}
\caption{Case study. The \textcolor[rgb]{0.7,0,0}{\underline{\textbf{\emph{underlined red}}}} words lead to inconsistent information, and the words with \colorbox[rgb]{0.8,0.8,0.8}{\color{black} gray} background are not informative enough. The \textcolor[rgb]{0,0.45,0.65}{\textbf{\emph{blue}}} words represent consistent and informative expressions.}\label{case_study}
\end{figure}

\subsection{Ablation Study}

To analyze which mechanisms are driving the improvements, we present an ablation study in Table \ref{tab:ablation}. We obtain ablated models including: 1) Pipeline, learning the two tasks sequentially; 2) DMTCI-M, only considering the mediator; 3) DMTCI-LDC, only considering the confounder. To avoid misleading, we denote our full model by DMTCI-Full. Table \ref{tab:ablation} shows the performance of these ablated models. We can see that Pipeline suffers from the train-test discrepancy and falls behind the baseline model TransLSTM. DMTCI-M increases the CIDEr-D score by $2.2$ and $2.6$ points over TransLSTM in MLE and RL periods, respectively. DMTCI-LDC outperforms TransLSTM while performing slightly worse than the DMTCI-M. When compared two ablated models, our full model achieves the best CIDEr-D score in both MLE and RL periods. And the selector further brings a large improvement in the MLE period and does not help much in the RL period.

\subsection{Case Study}\label{sec:case}

To help understand how the mediators and the confounders affect the learning, we show some cases in Figure \ref{case_study}. The inconsistent descriptions are marked by the red underlines and the not informative enough descriptions are marked by the gray background. The bold blue words represent applaudable expressions that are content consistent and informative. Compared with TransLSTM, DMTCI-M can alleviate the inconsistency problem by involving the mediator. When further combining with the latent de-confounding approach, DMTCI-Full generates consistent and informative captions.

Furthermore, we are trying to discuss why involving the mediators and the confounders work. At present, deep learning models have strong abilities to capture correlations. However, since it is too strong, models can often utilize the co-occurrence of concepts to minimize the training loss, ignoring the exact appearance of the detected objects. Then in the inference stage, it is possible to infer non-exist objects relying on the co-occurrence (e.g., "\emph{brush}" and "\emph{woman}"). Involving mediators and confounders during training could alleviate this problem by focusing on the exactly appeared objects.

\section{Conclusion}

In this paper, we propose a dependent multi-task framework with causal intervention for image captioning. We first involve BOC as the mediator to help the model understand the pre-extracted features, then we propose a latent de-confounding approach to alleviate the spurious correlation problem. Furthermore, we apply a multi-agent reinforcement learning strategy to better utilize the task dependencies and enable end-to-end training. The experiments show that our model yields competitive results with recent SOTA models and the proposed framework could improve content consistency and informativeness.

\section*{Acknowledgments}

The authors would like to thank the anonymous reviewers for their constructive comments. This work was supported by the National Key Research and Development Program of China under Grant 2018YFC0830400, and Shanghai Municipal Science and Technology Major Project under Grant 2021SHZDZX0102.

\appendix

\section{Gumbel Sampling}\label{apd_gs}

The Gumbel-softmax function is used when sampling $\widetilde{\boldsymbol{m}} \sim p(\boldsymbol{m} | \boldsymbol{R})$ which can back-propagate gradients between tasks. Denoting the one-hot representation of $\widetilde{\boldsymbol{m}}=\{\widetilde{m_j}\} \in \mathbb{R}^{L_m}$ by $\widetilde{\boldsymbol{Y}}^{m}=\{\widetilde{\boldsymbol{y}_j}\} \in \mathbb{R}^{L_m \times N_m}$ where $L_m$ denotes the number of all categories, and $N_m$ denotes the maximum number of detected objects belongs to each category, we assume that a logit value $\widetilde{\boldsymbol{o}}_j$ is obtained by a FCNN for $j$th category. A softmax function is used to produce the probability:
\begin{equation}
\widetilde{\boldsymbol{y}_j} = \text{softmax}(\boldsymbol{o}_j)
\end{equation}
Traditionally, we can sample a value $\widetilde{m}_j$ from $\widetilde{\boldsymbol{y}_j}$ with the multinomial function:
\begin{equation}
\widetilde{m_j} = \text{multinomial}(\widetilde{\boldsymbol{y}_j})
\end{equation}
which is non-differentiable. Gumbel-softmax uses a re-parameter trick by:
\begin{equation}\label{eq:gumbel}
\widehat{\boldsymbol{y}_j} = \text{softmax}((\boldsymbol{o}_j + \boldsymbol{g})/\tau)
\end{equation}
where $\boldsymbol{g}$ samples from $\text{Gumbel}(0, 1)$ and $\tau$ is the temperature. When $\tau \rightarrow 0$, $\widehat{\boldsymbol{y}_j}$ is approximated to the one-hot representation of a sampled value $\widetilde{m_j}$, which can be directly multiplied with embedding layers.

\section{Image-Text Selector}\label{apd_selector}

By sampling multiple latent variables $\boldsymbol{z}_c \sim p(\boldsymbol{z}_c)$, our model can generate multiple candidate captions $\widetilde{\boldsymbol{Y}} = (\widetilde{\boldsymbol{y}}^1, \widetilde{\boldsymbol{y}}^2, ..., \widetilde{\boldsymbol{y}}^{N_c})$ for an image $\boldsymbol{x}$ where $N_c$ is the number of generated captions. The remaining thing is to find out the best candidates by a trained selector, scoring each candidate caption by $s_i = S(\widetilde{\boldsymbol{y}}^i, \boldsymbol{x})$. It is worth noting that we can use the optimal selector in the training stage to score each candidate caption by $s^{\ast}_i = S^{\ast}(\widetilde{\boldsymbol{y}}^i, \boldsymbol{Y})$ where $\boldsymbol{Y}$ is the set of referenced captions, but not in the inference stage because $\boldsymbol{Y}$ is not available at that time.

Inspired by the previous work involving ranking candidate summaries in extractive summarization \cite{DBLP:conf/acl/ZhongLCWQH20}, and the previous work on the problem of image-text matching \cite{wei2020adversarial}, we use a candidate evaluator to select the best caption through a ranking objective.

We first adopt a Transformer \cite{vaswani_attention_2017} encoder followed with the average pooling layer to get the representation $\boldsymbol{r}_s$ for the image:
\begin{equation}\label{eq:disc_2}
\boldsymbol{H}_s = \text{Transformer}_{(s)}(\boldsymbol{R})
\end{equation}
\begin{equation}
\boldsymbol{r}_s = \text{pool}(\boldsymbol{H}_s)
\end{equation}
where the difference from the vanilla Transformer is that $\text{Transformer}_{(s)}(\cdot)$ does not contain the position embedding. Then we encode the generated captions $\widetilde{\boldsymbol{y}}^i$ through BERT \cite{devlin_bert:_2018}:
\begin{equation}
\boldsymbol{H}_s^i = \text{BERT}(\widetilde{\boldsymbol{y}}^i)
\end{equation}
\begin{equation}
\boldsymbol{h}_s^i = \text{pool}(\boldsymbol{H}_s^i)
\end{equation}
where $\boldsymbol{h}_s^i$ is the representation of $\widetilde{\boldsymbol{y}}^i$. Finally, we score the image-text pair $(\boldsymbol{x}, \widetilde{\boldsymbol{y}}^i)$ represented by $(\boldsymbol{r}_s, \boldsymbol{h}_s^i)$ as follows:
\begin{equation}
s(\widetilde{\boldsymbol{y}}^i, \boldsymbol{x}) = \sigma\left(\boldsymbol{W}_{s}\left[\boldsymbol{h}_s^i \oplus \boldsymbol{r}_s \oplus |\boldsymbol{h}_s^i-\boldsymbol{r}_s| \oplus \boldsymbol{h}_s^i \odot \boldsymbol{r}_s \ \right]\right)
\end{equation}
where $\boldsymbol{W}_{s}$ denotes the parameters of the scoring network and $\oplus$ denotes the concatenation operation. The score $s(\widetilde{\boldsymbol{y}}^i, \boldsymbol{x})$ is between $0$ and $1$, and better captions need to be closer to $1$. The scores of gold captions are set to $1$.

Then we use the margin-based triplet loss for the generated captions in two way: comparing with gold captions, and comparing between arbitrary two generated captions. Given $N_c$ generated candidate captions and $N_r$ referenced captions, we rank the generated captions in descending order based on the gold CIDEr-D \cite{DBLP:conf/cvpr/VedantamZP15} scores where $s^{\ast}_i = S^{\ast}(\widetilde{\boldsymbol{y}}^i, \boldsymbol{Y})$. The ranked captions are denoted by $\widetilde{\boldsymbol{Y}}_r = (\widetilde{\boldsymbol{y}}^1_r, \widetilde{\boldsymbol{y}}^2_r, ..., \widetilde{\boldsymbol{y}}^{N_c}_r)$ where $\widetilde{\boldsymbol{y}}^1_r$ has the highest score in $\widetilde{\boldsymbol{Y}}_r$. Then the loss is as follows:
\begin{equation}\label{candidate_selector}
\begin{aligned}
\mathcal{L}_s(\theta^s) &= \max \left(0, s(\widetilde{\boldsymbol{y}}^i_r, \boldsymbol{x})-s(\boldsymbol{y}^{\ast}, \boldsymbol{x}) + \gamma_{1}\right) 
\\&+ \max \left(0, s(\widetilde{\boldsymbol{y}}^j_r, \boldsymbol{x}) - s(\widetilde{\boldsymbol{y}}^i_r, \boldsymbol{x}) + \gamma_{2}\right)
\end{aligned}
\end{equation}
where $\gamma_{1}$ and $\gamma_{2}$ are the hyperparameters representing margin values, and $i$ and $j$ represent the ranked indexes. The first term is the triplet loss with gold captions and the second term is the triplet loss among generated captions. At the inference stage, we can select the best caption with the highest score in order to compare with state-of-the-art models.

\section{Discussion and Future Work}

As shown in Table 1 in the experiments section, without RL, training under MLE already produces high-quality candidate captions with the upper bound reaching a CIDEr-D score of $141.9$ points. We suggest that one possible future work is training a stronger image-text selector, taking the progress of cross-modal representations beyond the task of image captioning, such as visual question answering, visual reasoning, and image-text retrieval \cite{DBLP:conf/eccv/Li0LZHZWH0WCG20}. In this way, we may quickly train a generator without RL and fine-tune an image-text selector based on pre-trained cross-modal embedding networks. This paradigm may also bridge the gap between text generation and image/text understanding tasks.

% \small
\bibliographystyle{named}
\bibliography{ijcai2021.bib}

\end{document}